\documentclass{article}

\usepackage{microtype}
\usepackage{graphicx}
\usepackage{subfigure}
\usepackage{booktabs} 

\usepackage{amsmath}

\usepackage{float}

\usepackage{hyperref}

\usepackage{multirow}

\usepackage[accepted]{icml2020}

\icmltitlerunning{Quantized Neural Network Inference with Precision Batching}

\begin{document}
\setlength{\intextsep}{1pt}

\twocolumn[
\icmltitle{Quantized Neural Network Inference with Precision Batching}



\icmlsetsymbol{equal}{*}

\begin{icmlauthorlist}
\icmlauthor{Maximilian Lam}{}
\icmlauthor{Zachary Yedidia}{}
\icmlauthor{Colby Banbury}{}
\icmlauthor{Vijay Janapa Reddi}{}
\end{icmlauthorlist}

\icmlkeywords{quantization, performance, machine learning, RNN}

\vskip 0.3in
]




\begin{abstract}
 We present \textit{PrecisionBatching}, a quantized inference algorithm for speeding up neural network execution on traditional hardware platforms at low bitwidths without the need for retraining or recalibration. \textit{PrecisionBatching} decomposes a neural network into individual bitlayers and accumulates them using fast 1-bit operations while maintaining activations in full precision. \textit{PrecisionBatching} not only facilitates quantized inference at low bitwidths ($<8$ bits) without the need for retraining/recalibration, but also 1) enables traditional hardware platforms the ability to realize inference speedups at a finer granularity of quantization (e.g: 1-16 bit execution) and 2) allows accuracy and speedup tradeoffs at runtime by exposing the number of bitlayers to accumulate as a tunable parameter. Across a variety of applications (MNIST, language modeling, natural language inference) and neural network architectures (fully connected, RNN, LSTM), \textit{PrecisionBatching} yields end-to-end speedups of over $8 \times$ on a GPU within a $< 1\%$ error margin of the full precision baseline, outperforming traditional 8-bit quantized inference by over $1.5\times$-$2\times$ at the same error tolerance. 
\end{abstract}
\section{Introduction}
Recent advances in deep learning have demonstrated the wide range of the applications of neural networks \cite{cnn_imagenet, lstm, seq2seq, graves_speech, bengio_lm, rajpurkar-squad, wang_languageinfer, dao_learning}. Applying a neural network to make predictions, termed inference, is computationally expensive, exacts high energy costs and often demands specific latency requirements (e.g: a speech recognition neural network must be fast enough in decoding human language for a real time virtual assistant). Research in quantization aims to reduce the computational costs and thus improve the speed of neural network inference \cite{hubara_bnn, hubara_quant_1, xu_qrnn, sze_quant}. Generally, quantization reduces the precision of neural network weights / activations and speeds up inference by facilitating the use of high throughput low precision operations and by reducing the amount of memory transfers \cite{lin_fewmult, xu_qrnn, krishnan_quarl}. 
\\
\\
Importantly, quantization incurs an increasingly larger accuracy penalty when quantizing to lower bitwidths \cite{hubara_quant_1, pact}. For this reason, state of the art quantization methods often retrain or recalibrate their neural networks to achieve sufficient accuracy at lower bitwidths \cite{pact, zhu_ternary, xu_qrnn, han_learning, han_deep, lam_w2b, zhou_inq}. It is also common to require architectural changes to the neural network in addition to retraining \cite{zhu_ternary, xu_qrnn}. At the time of writing, state of the art quantization methods for bitwidths $<$ 8 typically involve retraining and recalibration. There are two key issues with this: 1) retraining a neural network for a target bitwidth is computationally expensive, often taking much longer to train than its full precision counterpart, requiring a separate hyperparameter tuning process for convergence and best results \cite{hubara_bnn, pact, hubara_quant_1} and 2) retraining/recalibrating a neural network requires access to data that matches the distribution of the training/test set \cite{nvidia_recalibrate, li_quantizetrain}, which may not be available (e.g: a machine-learning service provider like Google Cloud or Amazon Web Services may have access only to the model but not the data). Thus, a core motivation is the development of a quantized inference method that works out of the box and can extend to below 8 bits with corresponding speedups while maintaining accuracy. 
\\
\\
Additionally, research on quantized neural networks often involve the development of specialized hardware and it is unclear how their techniques translate to inference speedups on traditional hardware architectures (e.g: GPUs and CPUs) \cite{pact, bitpragmatic, sharma_bitpragmatic}. Concretely, traditional CPUs and GPUs lack the necessary datatypes for more unusual bitwidths (e.g: 2-bit, 3-bit). Thus, another core motivation is the development of a quantized inference algorithm that can be leveraged in context of existing hardware platforms to speed up inference at lower bitwidths ($<$ 8 bits).
\\
\\
We present \textit{PrecisionBatching} an ad-hoc algorithm for quantized inference targeted to traditional hardware platforms. \textit{PrecisionBatching} is based on the following observations:

\newenvironment{enumerate*}%
  {\begin{enumerate}%
    \setlength{\itemsep}{2pt}%
    \setlength{\parskip}{2pt}}%
  {\end{enumerate}}
  
  \newenvironment{itemize*}%
  {\begin{itemize}%
    \setlength{\itemsep}{2pt}%
    \setlength{\parskip}{2pt}}%
  {\end{itemize}}
 
\begin{figure}
  \centering
    \includegraphics[width=0.45\textwidth]{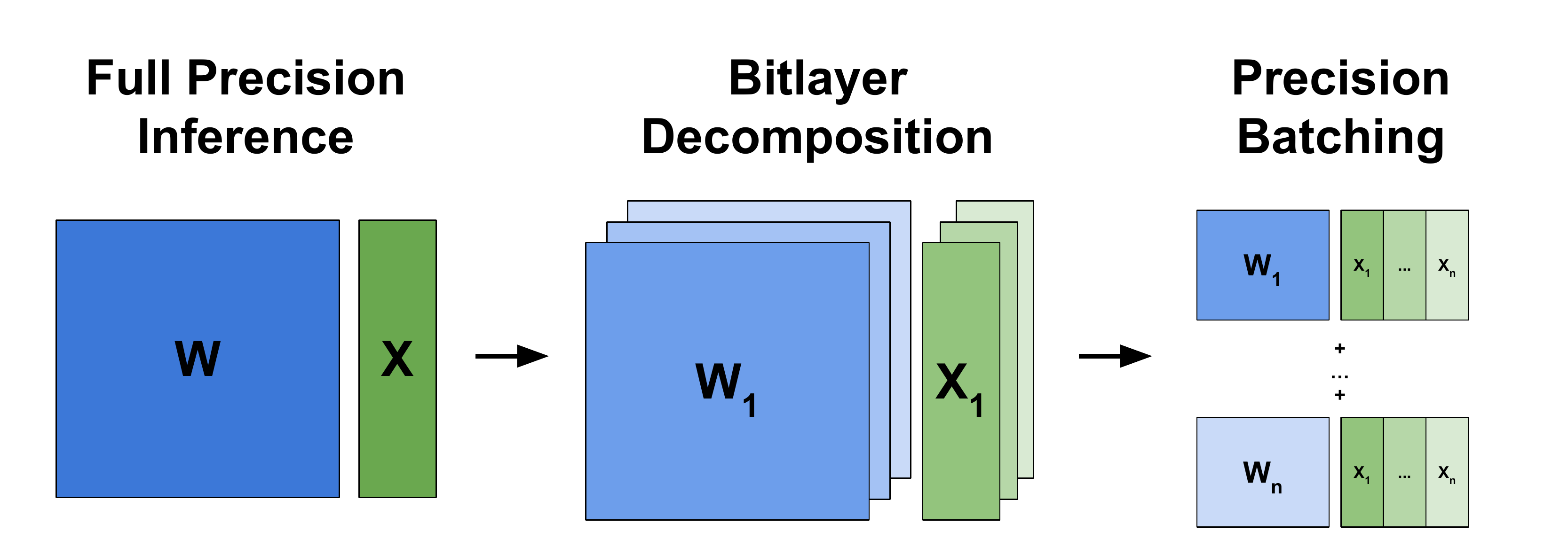}
  \caption{\textit{PrecisionBatching} quantized inference: 1) Decompose weights and activations into individual 1-bit tensors 2) Batch 1-bit activation tensors to enable efficient batched matrix multiplies against full precision activations 3) accumulate more terms to specify higher precision weights.}  
  \label{fig:pbatch}
   \vspace{-1.5em}
\end{figure}

\begin{enumerate*}
    \item Weights and activations may be decomposed into a sum of 1-bit tensors. Fast 1-bit operations may be leveraged for computation involving these terms. By accumulating more bitlayers, higher precision and accuracy may be attained at higher computational cost.
    \item Activations are harder to quantize than weights as they not only may have a larger spread (via multiply accumulate) but dynamically change depending on the input to the neural network. This motivates keeping activations in higher precision to avoid significant loss of accuracy and the need to retrain/recalibrate. Activations are kept in higher precision by batching the decomposed 1-bit tensors; results are obtained by reducing over the batch after the forward operation.
    \item Neural network inference in deployed applications typically involves small batch sizes indicating that computation is primarily limited by how fast data can be moved between the compute and memory units. Although batching across precision increases the amount of compute operations, this extra cost is hidden by memory accesses. An overall speedup is obtained by reducing the amount of memory accesses performed and is proportional to the number of bitlayers accumulated. 
\end{enumerate*}
The key idea of \textit{PrecisionBatching} is to leverage the high compute efficiency of traditional hardware platforms to operate over higher precision activations, leading to significant gains in model quality while maintaining the speed of low bitwidth quantized inference. By reframing $n$-bit weight, $k$-bit activation operations as a sum of 1-bit operations, any level of quantized execution (e.g: 1-16 bit) may be performed on traditional hardware platforms (Figure \ref{fig:pbatch}). As each product is performed over a binary matrix, memory is reduced by $32 \times$ per term, yielding a net speedup despite a $k \times$ increase in compute. We highlight that our algorithm is targeted for inference at deployment with a batch size of 1 to minimize latency \cite{fowers_realtime, 2015WhitepaperGD, han_eie}. 


To demonstrate the value of \textit{PrecisionBatching}, we develop optimized computational kernels to perform our algorithm on the GPU and evaluate our method against standard quantized inference implementations (NVIDIA's Cutlass linear algebra library \cite{nvidia_cutlass}) on various applications including fully connected networks for MNIST and LSTMs/RNNs for language modeling and natural language inference. Across this range of applications and models we demonstrate significant end-to-end speedups over using standard quantized inference methods ($>10 \times$ over full precision inference, $1.5\times$ - $2 \times$ over standard 8-bit quantized inference, at the same error margins). Our contributions are as follows
\begin{itemize*}
    \item We develop \textit{PrecisionBatching} an algorithm for quantized neural network inference targeted to traditional hardware platforms (e.g: CPU and GPU). \textit{PrecisionBatching} enables quantized inference at lower bitwidths and achieves better speedup per accuracy over standard quantized operations (e.g: 8/16-bit weights and activations operations) without retraining.
    \item We evaluate \textit{PrecisionBatching} over a variety of applications (MNIST, language modeling, natural language inference) and neural network architectures (fully connected, LSTM, RNN) and show net speedups of $> 10 \times$ over the full precision baseline ($> 1.5\times$-$2 \times $ over standard 8-bit quantized inference) within the same error tolerance. Furthermore, we leverage the finer granularity of precisions supported by \textit{PrecisionBatching} to boost speed vs model quality.
    \item We release optimized GPU kernels for our algorithm (and corresponding baselines) in the form of PyTorch modules.
\end{itemize*}

\section{Related Work}
\subsection{Post Training Quantization}
Post training quantization is the standard method for quantizing neural networks without retraining and involves clipping the values of a pre-trained model based on statistics \cite{outlier_channel}. Various methods for post training quantization have been researched. Naively, post training quantization involves casting weight and activation values to the nearest $n$-bit representation. More sophisticated techniques involve clipping the weights and activations so as to minimize some form of error between the quantized and real values \cite{nvidia_recalibrate, outlier_channel}. Even more advanced techniques change the underlying floating point format to enhance speed/accuracy \cite{tambe_adaptive, lee_log, johnson_float}.

Pre-existing research in post training quantization methods often omit details as to how the resulting quantized weights/activations may be leveraged on existing CPU and GPU platforms to speed up inference. More unusual bitwidths (e.g: 2/3/4/5) lack a corresponding data type on traditional hardware platforms and hence it is unclear how these levels of quantization improve inference. The implied benefit of post training quantization methods on these bitwidths is either space/memory savings or deployment to specially developed hardware accelerators for which fixed point operations for various bitwidths may be developed. By framing $n$-bit fixed point inference operations as a sum of binary operations, \textit{PrecisionBatching} is an effective solution to realize these quantization gains on traditional hardware platforms. Hence, \textit{PrecisionBatching} extends the memory-savings benefits of various post training quantization methods to speed gains on traditional hardware architectures.

\subsection{PACT}
The importance of activations in quantization quality has been noted in research. Specifically, PACT (Parameterized Clipping Activation for Quantized Neural Networks) \cite{pact} demonstrated that neural network weights and activations may be quantized to very low bitwidths ($< 4$) if an activation scale is optimized during training. Although PACT requires changes to the training process (and hence does not work out of the box), their research demonstrates the importance and difficulty of quantizing activations in maintaining quantization quality. Motivated by their findings, \textit{PrecisionBatching} opts to keep activations in higher precision (8,16,32 bit) to maintain accuracy at very low quantization level. This comes at minimal cost during inference as compute is dominated by memory access times. Thus, \textit{PrecisionBatching} circumvents the need to maintain a quantization scale at training time by giving more bits of precision to activations at inference time.

\subsection{Outlier Channel Splitting}
Recently, research into quantization without retraining has emerged as a topic of interest. One notable work is Outlier Channel Splitting \cite{outlier_channel}, which eliminates large magnitude weights/activations (which increase quantization error) by splitting them into separate channels, then applying standard post training quantization on the splitted weights, improving quantization performance. Outlier Channel Splitting demonstrates better performance-per-bit by using their technique in conjunction with standard post training quantization methods. Importantly, the authors note that outlier channel splitting may also be done to activations at runtime, though this is computationally difficult as it requires repeatedly finding the maximum of a matrix and adding rows to it. \textit{PrecisionBatching} eliminates this need by using more bits to represent activations, improving accuracy. Like many standard post-training quantization methods, Outlier Channel Splitting may be applied along with \textit{PrecisionBatching} to improve quantization quality and to extend their memory-saving gains to speed gains on traditional hardware platforms.

\subsection{Bitserial Computation}
In hardware architecture research, bitserial computation is a technique similar to \textit{PrecisionBatching} for quantized inference and similarly operates by decomposing fixed point operations into bitwise operations \cite{stripes, bitpragmatic, sharma_bitpragmatic}. Like \textit{PrecisionBatching}, bitserial computation frames $n$-bit fixed point operations as a sum of bitwise operations and accumulates the result layer by layer. Importantly, this technique/formulation is applied primarily to develop specialized hardware accelerators for machine learning and realizing the technique requires dedicated hardware constructs. Various hardware accelerators that leverage the bitserial formulation to reduce energy costs include \cite{stripes, bitpragmatic, sharma_bitpragmatic}. \textit{PrecisionBatching} differs as it targets traditional CPU and GPU platforms and does not require the development of specialized hardware. The key idea is that on traditional architectures low batched inference is memory bound and by batching the decomposed 1-bit vectors the extra overhead in compute is negated by the reduction in memory accesses, yielding a net speedup. 

\subsection{Streamlined Deployment for Quantized Neural Networks}
Another related work to \textit{PrecisionBatching} is Streamlined Deployment for Quantized Neural Networks \cite{streamlined_deployment}, which leverages a bitserial formulation to speed up deployment on the CPU. Similar to \textit{PrecisionBatching}, Streamlined Deployment for Quantized Neural Networks frames quantized operations in terms of 1-bit operations. However, the key difference is that Streamlined Deployment separates the the bitlayers of the activations into different product terms, rather than batching them into one large matrix multiplication. As shown in their paper, the impact is that both weights and activations must be kept in very low precision (e.g: 2-bit activations) due to the computational overhead of performing multiple matrix products, which naturally leads to significant degredation in accuracy. The key observation of \textit{PrecisionBatching} is that activation bitlayers may be batched together into one single matrix and a single large matrix product may be performed over this batch at high efficiency. This allows quantized inference with activations at or near full precision with minimal computational overhead, enhancing quantization performance.

\section{Precision Batching}
\subsection{Precision Batching Quantized Inference}
At a high level, \textit{PrecisionBatching} decomposes weights and activations into 1-bit tensors and replaces the main matrix-vector multiplication operation with a sum of 1-bit matrix-matrix operations. The core operation of neural network inference with a batch size of 1 is matrix-vector multiplication. 
$$
L_{i}(x) = Wx
$$
$L_{i}$ represents the function that transforms activation input $x$ at the specific layer of the neural network and $W$ is the trained weights of the neural network at layer $i$. Assuming that $W > 0$ and $x > 0$, we can decompose $W$ and $x$ into a sum of bitlayers (binary tensors) as in fixed point format
$$
W = \frac{1}{2^{16}}(2^{n-1} W^{(b)}_1 + ... + 2^{0} W^{(b)}_n) \mbox{ where } W^{(b)}_i \in [0, 1]
$$
$$
x = \frac{1}{2^{16}}(2^{k-1} x^{(b)}_1 +  ... + 2^{0} x^{(b)}_k) \mbox{ where } x^{(b)}_i \in [0,1]
$$
In the decomposition above, $n$ and $k$ represent the precision at which weights and activations are quantized to, respectively. Making $n$ and $k$ larger provides more accurate approximations of $W$ and $x$. $n$ describes the precision at which $W$ is estimated and represents the number of bitlayers to accumulate. The fraction $\frac{1}{2^{16}}$ represents the location of the fixed point and enables representation of values 16 binary digits $< 1$. The fixed point may be changed depending on the scale of values of the weights and activations.
Substituting back into the first equation and rearranging we get
$$
L_{i}(x) = Wx
$$
$$
= \frac{1}{2^{32}}(2^{n-1} W^{(b)}_1 +  ... + 2^{0} W^{(b)}_n)(2^{k} x^{(b)}_1 + ... + 2^{0} x^{(b)}_k)
$$
$$
= \frac{1}{2^{32}}\sum_{i=0}^{n} 2^{n-i} W^{(b)}_i (2^{k} x^{(b)}_1 + ... + 2^{0} x^{(b)}_k)
$$
The key observation is that the terms of the sum above can be rewritten as a single matrix multiplication. The idea is to batch together the bitlayer decomposition of $x$ into a single matrix and to frame the equation as a sum of matrix-matrix products.
$$
 \frac{1}{2^{32}} \sum_{i=0}^{n} 2^{n-i-1} (W^{(b)}_i [x^{(b)}_1 ... x^{(b)}_k]) [2^k ... 2^0]
$$
The main workload $W^{(b)}_i [x^{(b)}_1 ... x^{(b)}_k]$ exclusively consists of terms that are binary and facilitates efficient computation using 1-bit operations on CPU and GPU. Memory is reduced by a factor of approximately $\frac{32}{n}$, given that the matrix $W$ dominates the majority of memory accesses. Note that the number of compute ops is increased by a factor of $k$ as separating out the sum induces more work. However, as the reformulation leverages batching, the cost of the extra compute is negated by the higher computational efficiency of the matrix-matrix multiplication, and the reduction in memory accesses yields a net speedup. Concretely, memory overheads are often $10\times$-$100\times$ slower than a typical operation \cite{norvig} and significant gains may be attained by reducing these memory operations at the cost of extra arithmetic instructions.

As indicated, by choosing $n$ and $k$, any precision of weights and activations can be attained. In this paper $k$ (activation precision) is set to either 8, 16 or 32. Note that higher activation precision does not linearly impact performance due to the increase in computational efficiency. However, for CPUs that are less efficient (more compute bound), setting $k$ to be lower may significantly improve overall speed versus accuracy; hence $k$ and $n$ are parameters that determine the precision and speedup for quantized execution and may be tuned to the platform and requirement at hand. We analyze the impact of varying $n$ and $k$ on both speed and accuracy in the results.

Note that both the inputs and outputs of the \textit{PrecisionBatching} algorithm (as well as intermediate values such as partial sum accumulators) are full precision. The overhead of maintaining inputs and outputs as full precision is minimal as much of the computational and memory costs are attributed to large matrix multiply routines which are quantized (much of the memory costs are from loading the weights, rather than loading activations/inputs). Thus, keeping the intermediate inputs/activations in full precision is still aligned with the high level goal of speeding up inference.

Additionally, while the \textit{PrecisionBatching} formulation primarily targets matrix-vector multiplication (appropriate under the assumption that execution over a batch size of 1 is a matrix-vector multiply), this technique can be extended to any matrix-matrix product and hence handle any routine that involves general matrix multiplication. 

In this work, we primarily investigate models where inference over a batch size of 1 is handled by a matrix-vector multiplication, which limits us to feed forward networks, RNNs and LSTMs. Important future work involves the extension of this technique to general matrix products, which may be applied to CNNs, transformers and batched low precision training.

\subsection{Extending to Negative Values}
Note in the previous formulation that we assume $W > 0$ and $x > 0$. Here we extend the formulation to any real valued $W$ and $x$ matrix. Allowing any real valued input and matrix is important as it enables $PrecisionBatching$ to handle weights with negative values and cases where the input is not passed through a positive activation function (e.g: the first layer of the neural network whose inputs are real and may potentially contain negative values). The simple but effective idea is to leverage two's complement by adding an extra bitlayer with a negative scale to handle negative values. 
$$
W = \frac{1}{2^{16}}(-2^{n} W^{(b)}_0 + 2^{n-1} W^{(b)}_1 + ... + 2^{0} W^{(b)}_n) \mbox{, } W^{(b)}_i \in [0, 1]
$$
$$
x = \frac{1}{2^{16}}(-2^{k} x^{(b)}_0 + 2^{k-1} x^{(b)}_1 +  ... + 2^{0} x^{(b)}_k) \mbox{, }  x^{(b)}_i \in [0,1]
$$
Here, the first bitlayer for both $x$ and $W$ are negated, allowing for a complete representation of values between [$-2^{n}$, $2^{n}-1]$. This formulation is logically equivalent to two's complement format. Note that this technique incurs an extra bitlayer of computational overhead (for weights) and thus increases the computational and memory costs; we found in practice that the extra bitlayer of computational overhead for activations is minimal.

\subsection{Weight/Activation Quantization}
In the \textit{PrecisionBatching} formulation, $W$ and $x$ are effectively converted into fixed point format and quantized to reduce computation and memory accesses. However, any standard post training quantization technique (e.g: KL divergence,  MSE, etc) can be applied to $W$ and $x$ to improve accuracy, as long as the resulting set of quantization values are linearly spaced.

For applications, we use  standard post training quantization before quantized execution. 
$$
Q(W) = d \times round\left(\frac{W}{d}\right)\mbox{, } d = \frac{max(W)-min(W)}{2^n}
$$

Effectively, this rounds $W$ to the corresponding closest $n$-bit representable fixed point values. We found that in practice, rounding produces significantly better results than truncation at very lower bitwidths ($< 4$ bits). Additionally, for quantizing to $1$ bit, we found it extremely beneficial to exclude representing $0$ and instead opt to represent a positive and negative value. After the $n$-bit rounding, $Q(W)$ is applied in the \textit{PrecisionBatching} algorithm where the corresponding bitlayers and scales are deduced. Additionally, we also optimize over a clipping threshold to find a quantized matrix with the smallest mean error versus the full precision weight matrix. Note that quantizing $W$ is a preprocessing step that is done offline and hence does not affect inference performance measurements.

Quantizing activations $x$ utilizes a much simpler and efficient algorithm as it has to be done at runtime: truncation. For $x$ we simply convert $x$ from floating point to 32-bit fixed point (integer format) with a multiplication and a cast, which naturally drops bits outside the 32-bit range. 

The full \textit{PrecisionBatching} algorithm is broken into two stages: a preprocessing step which converts full precision weights to bitlayers, listed in algorithm \ref{alg:pbatch_prep}, and the inference stage which makes predictions given a full precision input, listed in algorithm \ref{alg:pbatch_infer}.

\newcommand{\MYDEF}[2][.8\linewidth]{%
  \leavevmode\hfill\makebox[#1][l]{~#2}}
\newcommand{\Desc}[2]{\State \makebox[2em][l]{#1}#2}
\begin{algorithm}[H]
\caption{PrecisionBatching Quantization Preprocessing}
\begin{algorithmic}
  \STATE {\bfseries Input} 
  \STATE \ \ \ \  $W$ \MYDEF{Full precision weight matrix}
  \STATE \ \ \ \  $n$ \MYDEF{Number of bits to quantize }
  \STATE {\bfseries Output} 
  \STATE \ \ \ \  $W^{(b)}$ \MYDEF{Bitlayers corresponding to quantized W}
  \STATE \ \ \ \  $S$ \MYDEF{Scales corresponding to quantized bitlayers}
\end{algorithmic}
\begin{algorithmic}[1]
  
  \STATE $W_q$ $\longleftarrow$ $Int(QuantizeRound(W, n) \times 2^{16})$
  \STATE $max\_bit$ $\longleftarrow$ $max$($log_2$($|W_q|$))
  \STATE $W^{(b)}$ $\longleftarrow$ [$W_q$ $\land$ $(1 \ll i)$ for $i$ in $max\_bit-n$ .. $max\_bit+1$]
  \STATE $S$ $\longleftarrow$ [1 $\ll$ $i$ for $i$ in $max\_bit-n$ .. $max\_bit+1$]
  \STATE $S[0]$ $\longleftarrow$ $S[0]$ $\times$ $-1$
  \STATE return $W^{(b)}$, $S$
  \end{algorithmic}
\label{alg:pbatch_prep}
\end{algorithm}

\begin{algorithm}[H]
\caption{PrecisionBatching Quantized Inference}
\begin{algorithmic}
  \STATE {\bfseries Input}  
  \STATE \ \ \ \   $W^{(b)}$ \MYDEF{Weight bitlayers}
  \STATE \ \ \ \   $S$ \MYDEF{Weight bitlayer scales }
  \STATE \ \ \ \   $x$ \MYDEF{Full precision input}
  \STATE {\bfseries Output} 
  \STATE \ \ \ \   $z$ \MYDEF{Full precision prediction} 
\end{algorithmic}
\begin{algorithmic}[1]
  \STATE $z \longleftarrow 0$
  \STATE $x_q \longleftarrow Int(x \times 2^{16})$
  \FOR{$W_{b}$, $scale$ in $W^{(b)}$, $S$}
    \STATE $z \longleftarrow z + \frac{scale}{2^{32}} \times (W_{b} x_q)[-2^{31} \mbox{ }2^{30} \mbox{ } .. \mbox{ } 2^0]$
  \ENDFOR 
  \STATE return $z$
\end{algorithmic}
\label{alg:pbatch_infer}
\end{algorithm}
\subsection{Efficient Implementation}
As indicated above, the core computation is an accumulation of products of binary tensors.
$$W^{(b)}_i [x^{(b)}_1 ... x^{(b)}_k]$$
As all values are 0 or 1, memory is reduced by packing the 0s and 1s into the bits of an integer array, yielding $32 \times$ reduction in memory for each product of bitlayers. Operating over these packed formats is inspired by standard binary quantized neural networks which uses logical operations and popcounts for implementing multiply accumulate. An important difference is that typical binary quantized neural network weights contain values that are -1 or 1 rather than 0 or 1. Hence, instead of the $xnor$ operation we use the $and$ operation to simulate 1-bit multiplication. To leverage these instructions, the floating point input vector must be converted to fixed point and then packed in such a way to layout the bits to be conducive to the $and/popcount$ instruction. Conversion to fixed point is a simple multiply and cast. Rearranging the bits is done with a bitwise matrix transpose, for which there are efficient implementations on both CPUs and GPUs that leverage parallelism / SIMD. In practice, we found the bitwise matrix transpose to have negligible overhead. We furthermore note that multiple bitlayers may be stacked together so that the entire product across bitlayers can be performed with a single operation. However, in practice we found that there is negligible performance difference in accumulating multiple bitlayers separately.

\subsection{Integer Quantized Inference}
Standard quantized inference methods quantize both weight and activation to the same precision before execution (so that both operands are the same datatype); for example, 8-bit quantized execution quantizes both weights and activations to 8-bit ints before operation. Weights and activations are scaled down before quantization (so that the maximum value is representable in the quantized range), then dequantized after the operation. Like in \textit{PrecisionBatching} we apply the same quantization preprocessing techniques (rounding, optimizing a clipping threshold) to weights before evaluation. Traditional CPU and GPU platforms provide support for 8-bit integer matrix operations and 16-bit floating point operations; more recent GPUs with tensorcores also support 4-bit and 1-bit integer matrix multiply operations. In our experiments, we leverage NVIDIA's T4 tensorcore capability (via NVIDIA's Cutlass linear algebra library) in the implementation of the standard quantized inference baselines.

\section{Results}
\subsection{Precision Batching Kernel Performance}
We implement optimized GPU kernels for the \textit{PrecisionBatching} algorithm and measure the speedup of the kernel over the full precision (32-bit) operation (provided by NVIDIA's Cutlass linear algebra library) across multiple precisions and matrix sizes. Inference times include all activation processing steps necessary for the algorithm, for example, transposing the activation bitmatrix before 1-bit execution. Baseline 4, 8 and 16 bit matrix multiplies utilize the NVIDIA Cutlass library which performs low-precision matrix multiply using WMMA (warp matrix multiply accumulate) hardware operations that leverage Tensorcores for compute. In all experiments the batch dimension is 1.

We perform all performance benchmarks on NVIDIA's Tesla T4 GPU. We measure the wall-clock time of performing 1000 iterations of the target algorithm (to amortize startup and cache costs). Note that performance gains on applications may be higher than those reported in kernel measurements as applications access more memory than in the benchmarks.

\begin{table}[t]                           
 \centering
    \resizebox{.48\textwidth}{!}{\begin{tabular}{|l|c|c|c|c|}
    \hline
    Method & 512x512 & 1024x1024 & 2048x2048 & 4096x4096\\
    \hline
    PBatch-1 (a=8) & 10.8  & 13.8 & 12.0 & 13.6\\
    PBatch-1 (a=16) & 9.5 & 12.1 & 10.3 & 13.2\\
    PBatch-1 (a=32) & 8.0 & 10.7 & 8.0 & 10.7\\
    PBatch-2 (a=8) & 6.6 & 9.9 & 8.3 & 11.8\\
    PBatch-2 (a=16) & 6.8 & 8.8 & 7.1 & 10.9\\
    PBatch-2 (a=32) &  5.7 & 7.5 & 5.4 & 8.3\\
    PBatch-4 (a=8) & 4.9 & 6.5 & 5.1 & 7.3\\
    PBatch-4 (a=16) & 4.2 & 5.5 & 4.3 & 6.8\\
    PBatch-4 (a=32) & 3.6 & 4.8 & 3.4 & 5.3\\
    PBatch-8 (a=8) & 2.9 & 3.6 & 3.2 & 4.7\\
    PBatch-8 (a=16) & 2.5 & 3.2 & 2.5 & 4.0\\
    PBatch-8 (a=32) & 2.0 & 2.7 & 2.1 & 3.1\\
    \hline
    Int1 & 3.6 & 5.0 & 8.5 & 34.3\\
    Int4 & 3.6 & 4.7 & 5.8 & 11.0\\
    Int8 & 3.3 & 4.0 & 4.2 & 8.0\\
    Float16 & 2.3 & 1.8 & 2.0 & 2.8\\
    Float32 & 1 & 1 & 1 & 1\\
    \hline
    \end{tabular}}
    \caption{Quantized inference speedups over 32-bit inference across different methods, matrix sizes and activation quantization levels on the NVIDIA T4 GPU. PBatch-n (a=k) means n+1 bitlayers are accumulated with k-bit activations. (n-bit weights, k-bit activations)}
    \label{tab:pbatch_perf}                            
\vspace{-1.5em}
\end{table}

Table \ref{tab:pbatch_perf} shows the performance of the  \textit{PrecisionBatching} kernel with weight bits $\in (1, 2, 4, 8)$ and activation bits $\in (8, 16, 32)$, along with baseline quantizated inference kernels (Int1, Int4, Int8, Float16, Float32). We see that at fewer bits, the \textit{PrecisionBatching} kernel achieves significant speedups over full precision inference: 10-14x speedup for 1-bit, 5-7x for 4-bit (note that the optimal speedup for PBatch-n is $\frac{32}{n+1}$ with the sign layer taken into account). Using fewer activation bits increases performance only slightly as compute is not the main bottleneck in these operations. 

Generally, higher performance is seen at larger matrix sizes as the effect of the reduction in memory on performance is more pronounced. Baseline kernels (Int1, Int4, Int8 especially) perform much better at larger matrix sizes; we believe this is the case as their kernels are more optimized than ours and leverage Tensorcore capability for more efficient compute. 

\subsection{Benefits of Higher Precision Activations}
Next we show that using higher precision for activations leads to significantly better model accuracy at low bitwidths. We benchmark model accuracy across three applications: MNIST, language modeling and natural language inference. For each we train one baseline full precision model and evaluate the effects of various levels of weight and activation quantization on the model's end performance. For each model/application we quantize weights and activations to 1, 4, 8, 16 and 32 bits.

For the MNIST task \cite{lecun_mnist}, we train a 3-layer fully connected neural network with a hidden size of 4096 for 20 epochs, reaching a baseline accuracy of 98\%. We uniformly quantize the weights and activations of each layer to the target precisions. 

For language modeling, we train a model with a 1-layer 2048 unit LSTM \cite{lstm} as the encoder, and a 1-layer 2048 unit fully connected as the decoder (a common architecture used in language modeling \cite{melis_lm}). We apply dropout with a factor of .5 to the inputs of the encoder LSTM's recurrence, and to the encoder LSTM's output. We train the model on the Wikitext-2 dataset \cite{merity_wikitext} for 40 epochs, reaching a baseline perplexity of ~93. During evaluation of quantization on model accuracy, we quantize the LSTM's input and hidden layers to the same weight and activation levels; however, we keep the final fully connected decoder in full precision. 

For natural language inference, we train a model with a 1-layer 3072 unit LSTM encoder and a 3-layer 3072 unit fully connected decoder (a larger version of that seen in \cite{snli}). We train on the SNLI dataset \cite{snli} for 10 epochs and reach a baseline accuracy of 78\%. During evaluation of quantization on model accuracy, we uniformly quantize both the weights and activations of the LSTM encoder and the fully connected decoder to the target precisions.

\begin{table}[t]                           
 \centering
    \resizebox{.48\textwidth}{!}{\begin{tabular}{|c|l|c|c|c|c|c|}
    \hline
    Task &  & $Q_{activ.}=32$ & $Q_{activ.}=16$ & $Q_{activ.}=8$ & $Q_{activ.}=Q_{weight}$\\
    \hline
    \multirow{ 5}{*}{MNIST (acc.)}
    & $Q_{weight}=1$ & 85.8 & 86.7 & \textbf{87} & 10.1 \\
    & $Q_{weight}=4$ & 97.1 & \textbf{97.3} & \textbf{97.3} & 94.3 \\ 
    & $Q_{weight}=8$ & \textbf{98.0} & 97.8 & 97.8 & \textbf{98.0} \\
    & $Q_{weight}=16$ & \textbf{98.0} & 97.9 & 97.9 & \textbf{98.0} \\    
    & $Q_{weight}=32$ & - & - & - & 98.0 \\
    \hline
    \multirow{ 5}{*}{Language Modeling (ppl.)}
    & $Q_{weight}=1$ & \textbf{188.0} & 188.0 & 188.0 & 828.1 \\
    & $Q_{weight}=4$ & \textbf{94.3} &  \textbf{94.3} &  \textbf{94.3} & 148.9 \\ 
    & $Q_{weight}=8$ & \textbf{94.0} & \textbf{94.0} & \textbf{94.0} & \textbf{94.0} \\
    & $Q_{weight}=16$ & \textbf{91.7} & \textbf{91.7} & \textbf{91.7} & 92.8 \\    
    & $Q_{weight}=32$ & - & - & - & 92.8 \\  
    \hline
    \multirow{ 5}{*}{Natural Language Inference (acc.)}
    & $Q_{weight}=1$ & \textbf{76.1} & \textbf{76.1} & 74.0 & 32.8 \\
    & $Q_{weight}=4$ & \textbf{78.7} & \textbf{78.7} & 76.8 & 77.4 \\ 
    & $Q_{weight}=8$ & 78.9 & 78.9 & 76.9 & \textbf{79.1} \\
    & $Q_{weight}=16$ & \textbf{78.9} & \textbf{78.9} & 76.9 & 78.8 \\    
    & $Q_{weight}=32$ & - & - & - & 78.8 \\  
    \hline    
        
    \end{tabular}}
    \caption{Benefits of using more precision for activations on model quality, evaluated on MNIST, language modeling (Wikitext-2) and natural language inference (SNLI). Generally, using higher level activations allows quantizing twice as many bits (e.g: from 8-bits to 4-bits) with little degradation of model accuracy. Note that for accuracy (acc), higher is better, whereas for perplexity (ppl) lower is better (the best score for each weight precision is bolded).}
    \label{tab:benefits}  
    \vspace{-1.5em}
    
\end{table}

Table \ref{tab:benefits} shows model performance (accuracy for MNIST and natural language inference, perplexity for language modeling) for different weight and activations precisions. For weight bitlevels $< 8$, keeping activations at higher precision (8, 16 or 32 bit) greatly increases model accuracy; generally, keeping activations at a higher precision allows quantizing twice as many bits, from 8-bits to 4-bits, without significant loss in model accuracy. For MNIST, with 1-bit weights, using higher precision activations is the difference between $85\%$ accuracy and random guessing (~$10\%$ accuracy); with 4-bit weights, higher precision activations maintains within $<1\%$ of the full precision model's performance. Similarly, for language modeling, with 1-bit weights, higher precision activations reduces perplexity from ~800 to ~180; for 4-bit weights, higher precision activations reduce perplexity from ~180 to within a few points of the full precision performance. For natural language inference, using full precision activations allows us to quantize down to 1-bit with only a couple percentages of accuracy degredation (78\% to 76\%), whereas quantizing activations to 1-bit degrades to random guessing (33\%). Interestingly, for language inference, the 8-bit quantized model outperformed the full precision result, a known phenomenon seen in quantization literature \cite{krishnan_quarl, xu_qrnn}.

\subsection{End to End Performance Gains}

\begin{figure*}[!htp]
  \centering
  \subfigure[
  MNIST (FC) (higher accuracy is better).
  ]{\includegraphics[scale=0.33]{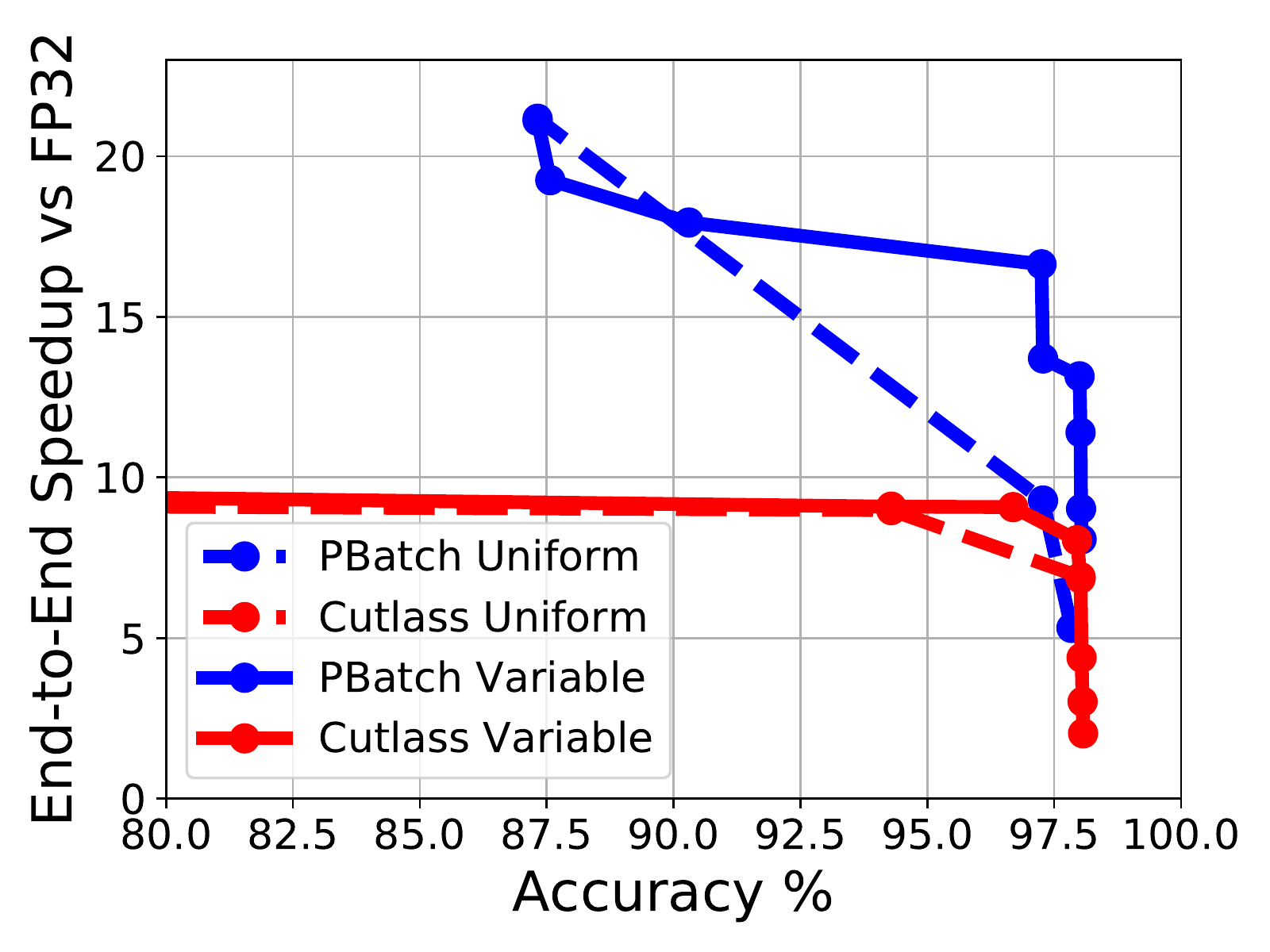}}\hfill
  \subfigure[Language modeling (LSTM) (lower perplexity is better).]{\includegraphics[scale=0.33]{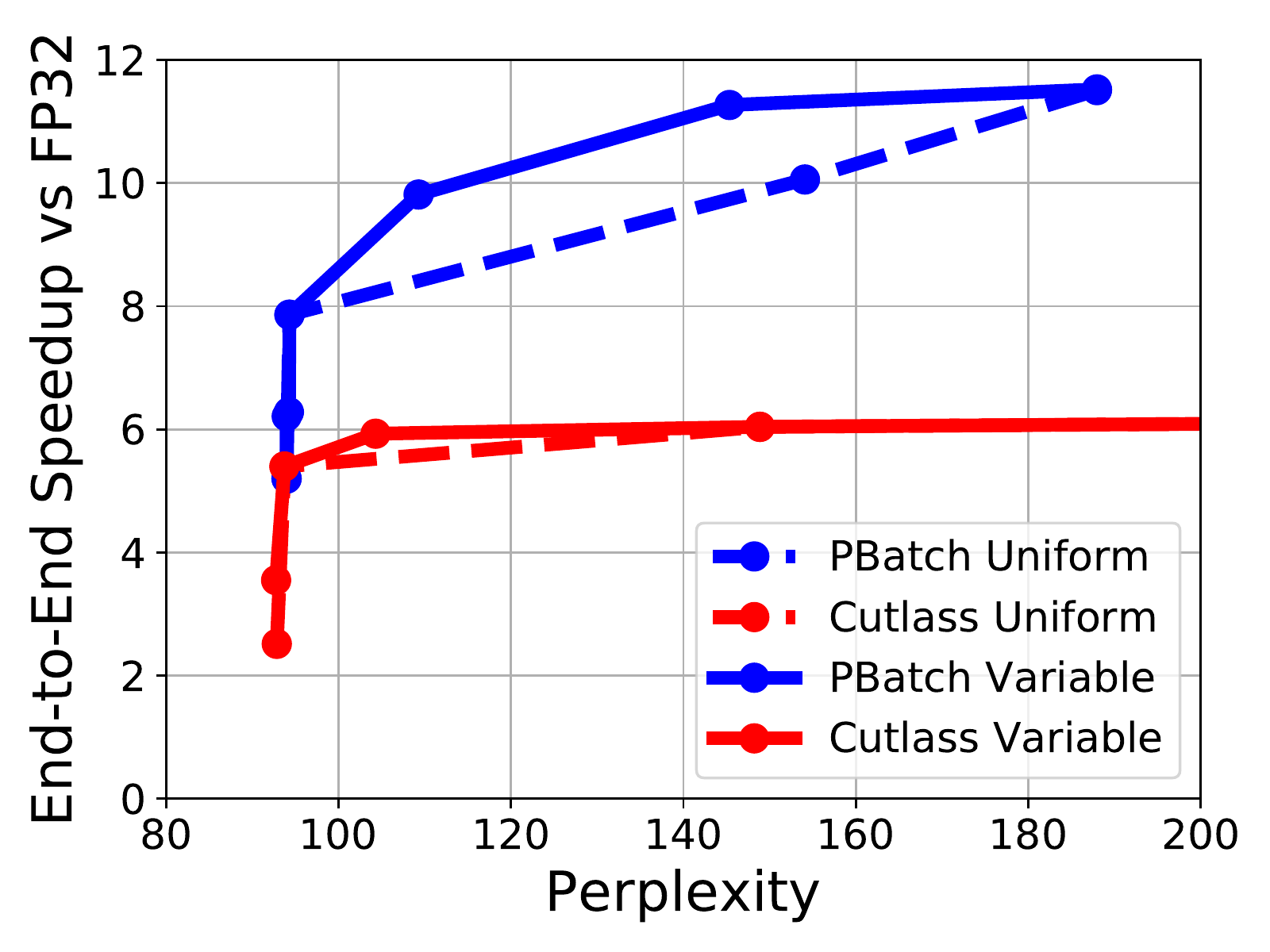}}\hfill
  \subfigure[Natural language inference (LSTM)  (higher accuracy is better).]{\includegraphics[scale=0.33]{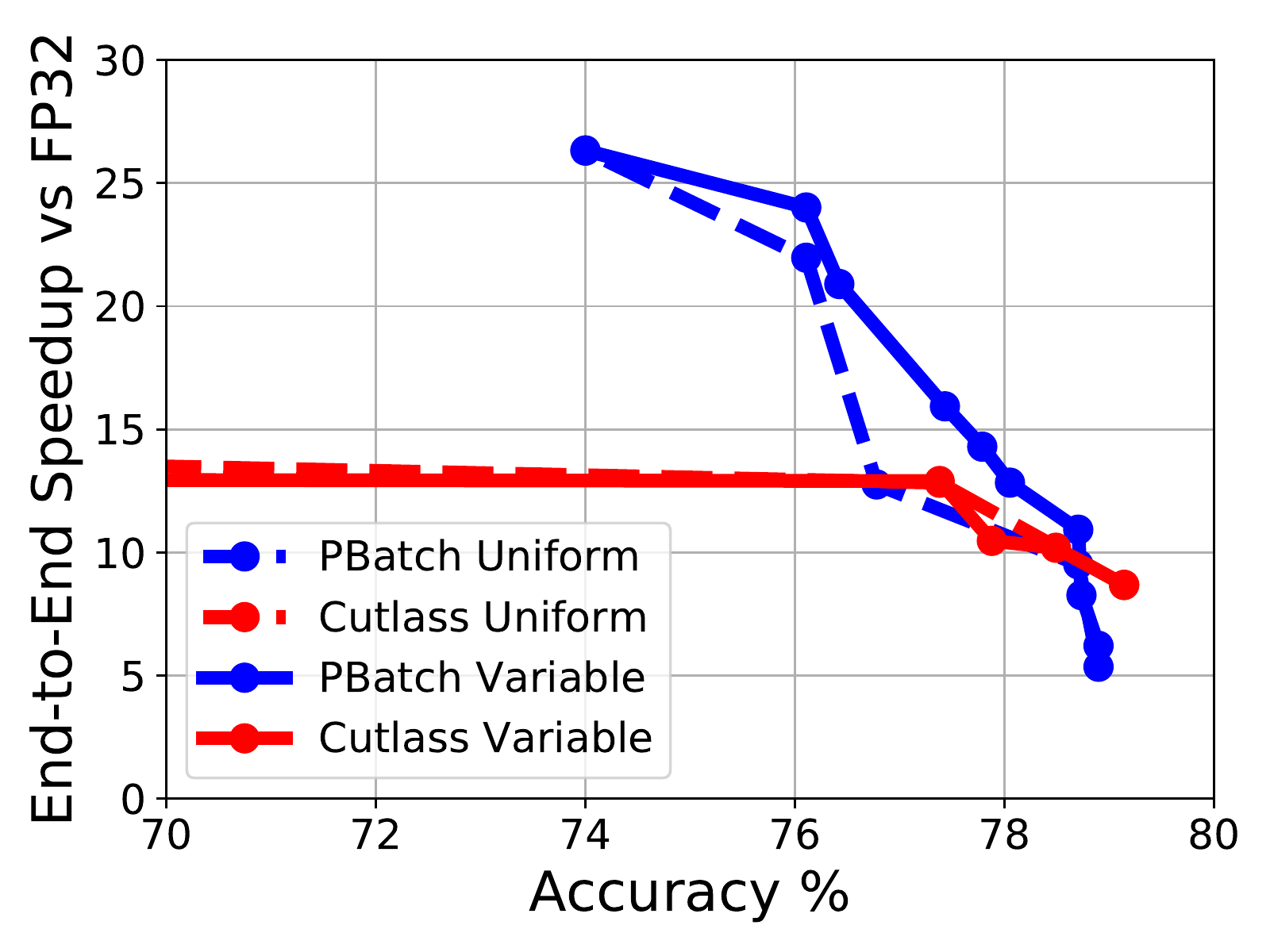}}  

\caption{End-to-end speedup over full precision model vs model quality on MNIST, language modeling and natural language inference over various precisions of weights (1,4,8,16,32 for baseline (Cutlass); 1,2,4,8 for PBatch) and activations (not applicable for baseline; 8,16,32 for PBatch). Dotted lines show Pareto boundary where all layers have the same precision (uniform layer quantization), while solid lines show Pareto boundary where layers may have different precision (variable layer quantization). \textit{PrecisionBatching} yields end-to-end speedups over $8 \times$ that of full precision inference, or $1.5 \times$ - $2\times$ over standard 8-bit quantized inference (Cutlass). Variable layer precision assignments perform noticeably better than uniform precision across weight layers, especially for \textit{PrecisionBatching}.}
\label{fig:endtoend}
\vspace{-1.0em}
\end{figure*}

We combine the observations from our previous results: we leverage the high runtime performance of the \textit{PrecisionBatching} kernel and the better model accuracy of keeping activations in higher precision to attain significant end-to-end speedups over the full precision model while maintaining model quality. We use the same applications (MNIST, language modeling (Wikitext-2) and natural language inference (SNLI)) with the same model architectures and training parameters described previously.

We apply each target quantized inference algorithm as follows. For the MNIST model, we replace each linear layer with the corresponding quantized inference algorithm; for the language modeling and natural language inference Seq2Seq model, we replace each linear layer of the encoder 1-layer LSTM with the target quantized inference algorithm, however we keep the final fully connected decoder in full precision.

Additionally, for both the baseline quantized inference and \textit{PrecisionBatching}, we use variable-bit quantization on different layers (e.g: 1-bit quantization on layer 1, 4-bit quantization on layer 2, etc) to further boost performance per accuracy. Accordingly, we perform an exhaustive grid search over weight/activation precision assignments. On the 3-layer fully connected for MNIST, for baseline quantized inference we assign each layer a precision $\in (1, 4, 8, 16, 32)$ (note that for quantized inference activations are the same precision as weights); for \textit{PrecisionBatching}, we assign each layer a precision $\in (1,2,3,4,8)$ and activations $\in (8, 16, 32)$. On the Seq2Seq LSTM for language modeling and natural language inference, for baseline quantized inference we assign each layer a precision $\in (1, 4, 8, 16)$; for \textit{PrecisionBatching}, we assign each layer a precision $\in (1,2,4,8)$ and activations $\in (8, 16, 32)$. 

In benchmarking the runtime performance of each model/application, we measure the wall clock time of inference with a batch size of 1 for 10 iterations on a given input repeated over 10 runs and take the minimum. We measure speedups by comparing the model with the target quantized inference algorithm against the model with the baseline quantized inference method.

Figure \ref{fig:endtoend} shows the Pareto curves of the end-to-end speedups of \textit{PrecisionBatching} over standard quantized inference for both uniform layer quantization (all layers the same precision) and variable layer quantization (different layers have different precisions). On average, \textit{PrecisionBatching} yields speedups of $8 \times$ - $10 \times$ over full precision inference, and $1.5 \times$ - $2 \times$  over standard 8-bit quantized inference at the same error tolerance. Additionally, the finer granularity of precision supported by \textit{PrecisionBatching} enables greater speedup per accuracy when using variable quantization across layers. The same data is reflected in table \ref{tab:endtoend}, which shows the corresponding best achieved speedup for each method for different error margins.

\begin{table}[h]                           
 \centering
    \resizebox{.48\textwidth}{!}{\begin{tabular}{|p{1.3cm}|c|c|c|c|c|c|}
    \hline
    Task &  Error & Quality & Speedup vs FP32 & Speedup vs Int8 & Method & Precision Assign.\\
    \hline
    \multirow{ 6}{*}{\shortstack{MNIST \\ (acc.)}}
    & \multirow{ 2}{*}{$<1\%$}
       & 97.3\% & 16.6 & 2.4 & PBatch & (4,8)(1,8)(1,8)\\    
       & & 97.9\% & 8.0 & 1.2 & Baseline & (8,4,8)\\ \cline{3-7} 
    & \multirow{ 2}{*}{$<5\%$}
       & 97.3\% & 16.6 & 2.4 & PBatch & (4,8)(1,8)(1,8)\\
       & & 94.3\% & 9.1 & 1.3 & Baseline & (4,4,4)\\\cline{3-7}      
    & \multirow{ 2}{*}{$<15\%$}
       & 87.3\% & 21.0 & 3.1 & PBatch & (1,8)(1,8)(1,8)\\
       & & 94.3\% & 9.1 & 1.3 & Baseline & (4,4,4)\\ \cline{3-7}  
    \hline
    \multirow{ 6}{*}{\shortstack{Language \\ Modeling \\ (ppl.)}}
    & \multirow{ 2}{*}{$<5$}
       & 94.3 & 7.9 & 1.5 & PBatch & (4,8)(4,8)\\
       & & 93.7 & 5.4 & 1 & Baseline & (8,8)\\ \cline{3-7} 
    & \multirow{ 2}{*}{$<25$}
       & 109.3 & 9.8 & 1.8 & PBatch & (1,8)(4,8)\\
       & & 104.3 & 5.9 & 1.1 & Baseline & (4)(8)\\ \cline{3-7}      
    & \multirow{ 2}{*}{$<50$}
       & 145.3 & 11.3 & 2.1 & PBatch & (1,8)(2,8)\\
       & & 148.9 & 6.0 & 1.2 & Baseline & (4,4)\\ \cline{3-7}      
    \hline
    \multirow{ 6}{*}{\shortstack{Natural \\ Language \\ Inference \\ (acc.)}}
    & \multirow{ 2}{*}{$<1\%$}
       & 77.8 & 14.3 & 1.6 & PBatch & (4,16)(1,8)\\
       & & 77.9 & 10.5 & 1.2 & Baseline & (4,8)\\ \cline{3-7} 
    & \multirow{ 2}{*}{$<5\%$}
       & 74.0 & 26.3 & 3.0 & PBatch & (1,8)(1,8)\\
       & & 77.4 & 12.9 & 1.5 & Baseline & (4,4)\\  \cline{3-7}      
    & \multirow{ 2}{*}{$<15\%$}
       & 74.0 & 26.3 & 3.0 & PBatch & (1,8)(1,8)\\
       & & 77.4 & 12.9 & 1.5 & Baseline & (4,4)\\  \cline{3-7}      
    \hline    
    \end{tabular}}
    \caption{Best model quality, speedups and precision assignments per error margin for \textit{PrecisionBatching} and baseline quantized inference. \textit{PrecisionBatching} achieves $1.5\times$ -- $2\times$ the performance of the baseline within the same error margin. \textit{PrecisionBatching} precision assignments are of the form ($L_i$ bits, $A_i$ bits); quantized inference precision assignments are of the form ($L_i$=$A_i$ bits).}
    \label{tab:endtoend}  
\end{table}
\section{Discussion}
We present \textit{PrecisionBatching}, a quantized inference algorithm for speeding up neural network execution on traditional hardware platforms at low bitwidths without the need for retraining or recalibration. Across various models (fully connected, LSTMs, RNNs) and applications (MNIST, language modeling, natural language inference) we show that \textit{PrecisionBatching}  yields end-to-end speedups of over $8 \times$ that of full precision inference ($1.5\times$ -- $2\times$ that of standard 8-bit quantized inference) at the same error tolerance. 

Importantly, we see this work as a modest yet significant step towards tackling the broad and long-standing challenge of maintaining high system efficiency during neural network execution. In many areas of neural network execution performance is bottlenecked by the massive amount of memory transfers necessary for the task. \textit{PrecisionBatching} demonstrates that more precision for activations may be attained at minimal cost by leveraging the higher compute efficiency of traditional hardware platforms, leading to significant gains in model accuracy at low bitwidths. While this work demonstrates the gains on a GPU, important future work involves engineering this technique to peak performance on CPUs. Although the authors have made significant efforts to attain the same speedups on the CPU, the lack of vectorized popcount instructions on current hardware limited success; we believe future hardware with these capabilities (e.g: Intel's Ice Lake Processor) will facilitate significant performance gains on CPUs. Additionally, while this work focuses on matrix-vector multiplication (low batch dimension), it is extensible to general matrix-matrix products; important future work involves applying \textit{PrecisionBatching} to a broader range of models such as CNNs and attention-models as well as extending it to a training and federated learning setting.

\bibliography{pbatch.bib}
\bibliographystyle{icml2020}

\end{document}


\icmltitlerunning{Supplementary Material}
\onecolumn

\icmltitle{Supplementary Material}

\section{Detailed Example of Precision Batching}

Suppose we have
$$
Wx
$$
where
$$
W = \begin{bmatrix}
    1 & -1 \\
    4 & -4 \\
\end{bmatrix},
x = \begin{bmatrix}
1 \\
-2 
\end{bmatrix}
$$
Viewed in binary (in two's complement fixed point format)
$$
W = \begin{bmatrix}
0b0001 & 0b1111 \\
0b0100 & 0b1100\\
\end{bmatrix},
x = \begin{bmatrix}
0b0001 \\
0b1110 
\end{bmatrix}
$$
Decompose and extract scales from $W$ (preprocessing step done offline)

$$
W_1^{(b)} = \begin{bmatrix}
0 & 1\\
0 & 1\\
\end{bmatrix},
W_2^{(b)} = \begin{bmatrix}
0 & 1\\
1 & 1\\
\end{bmatrix},
W_3^{(b)} = \begin{bmatrix}
0 & 1\\
0 & 0\\
\end{bmatrix},
W_4^{(b)} = \begin{bmatrix}
1 & 1\\
0 & 0\\
\end{bmatrix}
$$
$$
S_1 = -2^3,
S_2 = 2^2,
S_3 = 2^1,
S_4 = 2^0,
$$
Verify that
$$
W = W_1^{(b)} S_1 + W_2^{(b)} S_2 + W_3^{(b)} S_3 + W_4^{(b)} S_4 
$$

$x$ is converted from floating point format to fixed point format with a cast. Note that in the real computational operation, as popcounts operate over adjacent bits, a bitwise matrix transpose is performed on $x$ during the fixed point cast conversion (this bitwise transpose step is ommitted in this example).
$$
x^{(b)} = \begin{bmatrix}
0 & 0 & 0 & 1\\
1 & 1 & 1 & 0\\
\end{bmatrix}
$$
$$
S_x = \begin{bmatrix}
-2^3 \\ 2^2 \\ 2^1 \\ 2^0
\end{bmatrix}
$$
Verify that
$$
x = x^{(b)} S_x
$$

Thus we have
$$
Wx = [\sum_{i=1}^{4} S_i W_i^{(b)} x^{(b)}] S_x
$$

Note that in this example only integer values were used for simplicity. With fractional values, the scaling factors would extend to $< 1$, allowing representation of values between 0 and 1.
\section{Extended Data on End-to-End Speedups}
The following is the raw data used to generate the final end-to-end speedup graphs. 

Tables have 3 sections
\begin{itemize}
    \item Model with uniform layer quantization using PrecisionBatching and Baseline
    \item PrecisionBatching with variable layer quantization (sorted by speedup)
    \item Baseline (Cutlass) with variable layer quantization (sorted by speedup)
\end{itemize}

For PrecisionBatching, model string in the form $(L_i \mbox{ bits},A_i \mbox{ bits})$.
\\
For baseline, model string in the form $(L_i \mbox{  bits}=A_i \mbox{ bits})$.

\subsection{Extended Data of End-to-End Speedups on MNIST}

\begin{longtable}{|c|c|c|c|c|c|}
    \hline
    Method & Acc & Inference Time & Speedup Over 32-bit & Speedup Over 8-bit & Model String\\
    \hline
    Cutlass-32 Uniform & 98.0002 & 0.184862 & 1 & 0.1449080936 & \\
Cutlass-16 Uniform & 98.010199 & 0.06254 & 2.955900224 & 0.4283338663 & \\
Cutlass-8 Uniform & 98.020198 & 0.026788 & 6.900925788 & 1 & \\
Cutlass-4 Uniform & 94.290571 & 0.020576 & 8.9843507 & 1.301905132 & \\
Cutlass-1 Uniform & 10.088991 & 0.019268 & 9.594249533 & 1.390284409 & \\
PBatch-1 (a=16) Uniform & 86.661334 & 0.009509 & 19.44074035 & 2.817120623 & \\
PBatch-2 (a=16) Uniform & 86.871313 & 0.013579 & 13.61381545 & 1.972752044 & \\
\hline
    \end{longtable}

\subsection{Extended Data of End-to-End Speedups on Language Modeling}

    \begin{longtable}{|c|c|c|c|c|c|}
    \hline
    Method & Ppl & Inference Time & Speedup Over Uniform 32-bit & Speedup Over 8-bit & Model String\\
    \hline
    Cutlass-32 Uniform & 92.841564 & 0.844581 & 1 & 0.1853096387 & \\
Cutlass-16 Uniform & 92.840852 & 0.335876 & 2.514561922 & 0.4659725613 & \\
Cutlass-8 Uniform & 93.735784 & 0.156509 & 5.396373371 & 1 & \\
Cutlass-4 Uniform & 148.874395 & 0.139829 & 6.040098978 & 1.11928856 & \\
Cutlass-1 Uniform & 828.141487 & 0.126426 & 6.680437568 & 1.237949472 & \\
PBatch-1 (a=16) Uniform & 187.958778 & 0.078339 & 10.7811052 & 1.997842709 & \\
PBatch-2 (a=16) Uniform & 154.092182 & 0.091756 & 9.204640568 & 1.705708619 & \\
\hline
    \end{longtable}

\subsection{Extended Data of End-to-End Speedups on Natural Language Inference}

    \begin{longtable}{|c|c|c|c|c|c|}
    \hline
    Method & Accuracy & Inference Time & Speedup Over Uniform 32-bit & Speedup Over 8-bit & Model String\\
    \hline
    Cutlass-32 Uniform & 78.817182 & 3.196121 & 1 & 0.1152315573 & \\
Cutlass-16 Uniform & 78.84772 & 1.182457 & 2.702949029 & 0.3114650258 & \\
Cutlass-8 Uniform & 79.142915 & 0.368294 & 8.678178303 & 1 & \\
Cutlass-4 Uniform & 77.381922 & 0.249101 & 12.83062292 & 1.47849266 & \\
Cutlass-1 Uniform & 32.766694 & 0.189934 & 16.82753483 & 1.939063043 & \\
PBatch-1 (a=16) Uniform & 76.109528 & 0.14549 & 21.96797718 & 2.53140422 & \\
PBatch-2 (a=16) Uniform & 59.517508 & 0.210884 & 15.155825 & 1.746429317 & \\
\hline
    \end{longtable}